\pdfoutput=1

\documentclass[11pt]{article}

\usepackage[]{EACL2024}

\usepackage{times}
\usepackage{latexsym}

\usepackage[T1]{fontenc}

\usepackage[utf8]{inputenc}

\usepackage{microtype}

\usepackage{inconsolata}

\usepackage{paralist}
\usepackage{multirow}
\usepackage[export]{adjustbox}
\usepackage{booktabs}
\usepackage{nicematrix}
\usepackage{xcolor}
\usepackage{colortbl}
\usepackage[outline]{contour}
\usepackage{enumitem}
\usepackage{siunitx}
\usepackage{todonotes}
\usepackage{subcaption}
\usepackage{graphicx}

\definecolor{lightred}{HTML}{f4dee0} 
\definecolor{lightgreen}{HTML}{ecf4eb}

\usepackage{colortbl}
\usepackage{placeins}
\usepackage{enumitem}
\usepackage{paralist}

\newcommand{\abr}[1]{\textsc{#1}}

\graphicspath{ {./figs/} }

\usepackage{tabularx}
\usepackage{soul}

\newcommand{\ku}{$^1$}
\newcommand{\rff}{$^2$}
\newcommand{\wu}{$^3$}

\title{The Role of Data Curation in Image Captioning}

\author{Wenyan Li\ku, Jonas F. Lotz\ku$^{,}$\rff, Chen Qiu\wu,  Desmond Elliott\ku \\
{\ku}Department of Computer Science, University of Copenhagen \\
{\rff}ROCKWOOL Foundation Research Unit \\
{\wu}School of Computer Science and Technology, Wuhan University of Science and Technology \\
\texttt{\{weli,jonasf.lotz,de\}@di.ku.dk, chen@wust.edu.cn}}

\begin{document}
\maketitle

\begin{abstract}
    
    Image captioning models are typically trained by treating all samples equally, neglecting to account for mismatched or otherwise difficult data points. In contrast, recent work has shown the effectiveness of training models by scheduling the data using curriculum learning strategies. This paper contributes to this direction by actively curating difficult samples in datasets \emph{without} increasing the total number of samples. 
    We explore the effect of using three data curation methods within the training process: complete removal of a sample, caption replacement, or image replacement via a text-to-image generation model. Experiments on the Flickr30K and COCO datasets with the BLIP and BEiT-3 models demonstrate that these curation methods do indeed yield improved image captioning models, underscoring their efficacy.

\end{abstract}

\section{Introduction}
\label{sec:intro}

    Image captioning is the task of generating grammatically correct and accurate descriptions of visual data, which involves understanding the identity of salient objects and their relationships~\cite{bernardi2016automatic,baltruvsaitis2018multimodal}. While existing models have made significant progress on this problem, there remains an inherent challenge: how to address the variations in learning difficulty that arise from diverse image-caption pairs~\citep{sharma-etal-2018-conceptual,Schuhmann2021LAION400MOD}.

    Image captioning models are usually trained by treating the entire training dataset equally, which overlooks the variations in the complexity of each data point. One attempt at addressing this issue has been to apply data filtering as a preprocessing stage to large-scale datasets to remove noisy data from the pretraining process~\citep{li2022blip, nguyen2023improving}. Several other image captioning techniques have relied on curriculum learning strategies \citep{bengio2009curriculum}, which schedule the training data with increased levels of complexity, effectively adapting the learning process to the difficulty of the task \citep{liu-etal-2021-competence, dong2021CL,zhang-etal-2022-cross, Alsharid21, ayyubi-etal-2023-learning}. In this paper, we aim to answer a fundamental question: can image captioning models be improved by not only recognizing variations in the data but also actively curating difficult samples?
    
    We introduce three data curation methods, each with the aim of improving the learning process while preserving the overall size of the training dataset. These methods include the complete removal of a sample, the replacement of captions, or the substitution of images using a text-to-image generation model. The targets of these methods are image-caption training samples that have unusually high losses with respect to the rest of the training dataset under the current model parameters. In other words, our approach focuses on the samples that are proving \emph{difficult} to model \citep{bengio2009curriculum, kumar2010self}. 
    
    The main findings of this paper are:
    \begin{itemize}
        \item Dynamic data curation enhances image captioning performance. The best strategy varies between datasets but is generalizable to different vision-language models.\footnote{We release the code for our curation framework at \url{https://github.com/lyan62/data-curation/}} 
        \vspace{-5pt}
        \item The extent of curation is a critical factor and dataset dependent. We find that curating more than 50\% of data negatively impacts the effectiveness of data curation.\vspace{-5pt}
        \item Image generation-based curation has potential benefits with specific techniques, but its potential benefit is limited by generation errors identified through a human study, which are not apparent from automatic evaluation metrics, such as CLIPScore  \citep{hessel2021clipscore}.
    \end{itemize}

\section{Related work}
\label{sec:related}
    \paragraph*{Data Curation in NLP} While still under-explored for image captioning, \citet{Rogers2021ChangingTW} highlighted the importance of data curation for deep learning and NLP. Several studies have adopted data curation for large language models: \citet{chen2023lingua} developed a general text curation framework based on large language models; \citet{pmlr-v162-kandpal22a} and \citet{lee-etal-2022-deduplicating} discussed the impact of deduplication for training; \citet{chang2023data} shows that careful curation alone can stabilize in-context learning.

    \vspace{-1mm}
    \paragraph{Image Captioning and Learning Strategies}
    Curriculum learning~\citep{bengio2009curriculum} and self-paced learning~\citep{kumar2010self} are techniques that adjust the learning process based on variations in the learning samples, leveraging loss values to estimate model competence. For image captioning, several studies have introduced diverse learning techniques aimed at customizing the model training process in terms of sample difficulty, incorporating both textual and visual features~\citep{Alsharid21, dong2021CL, zhang-etal-2022-cross}. Whereas these methods adjust model training using sorted data, our approach proposes an innovative perspective: adjusting training by curating data samples that exhibit outlier losses, while preserving the overall dataset size.

    \vspace{-1mm}
    \paragraph*{Text-to-image Generative Models} Text-to-image generative models, including diffusion models~\citep{song2021iclr, nichol2021icml}, have rapidly gained popularity and proven powerful. 
    Although recent large-scale latent diffusion models excel in generating high-resolution images with artistic and photo-realistic qualities~\citep{sd, nichol2022glide, dalle2, saharia2022photorealistic}, their application in multimodal tasks remains unexplored. Concurrently to our work, ~\citet{azizi2023synthetic} and~\citet{jain2023distilling} show that image classifiers can be improved by learning from augmented images generated by finetuned generative models; \citet{xiao2023multimodal} and \citet{caffagni2023synthcap} used generative models to augment the datasets used to train captioning models. %

   To the best of our knowledge, we are the first to explore how dynamic data curation approaches can impact downstream image captioning \emph{without} scaling up existing datasets, and how text-to-image generative models can be applied in the process.

\section{Data Curation for Captioning}
\label{sec:al}

    \begin{figure*}[!t]
        \centering
        \includegraphics[max size={0.9\linewidth}{0.8\textheight},trim={2mm 0mm 5mm 3mm}, clip]{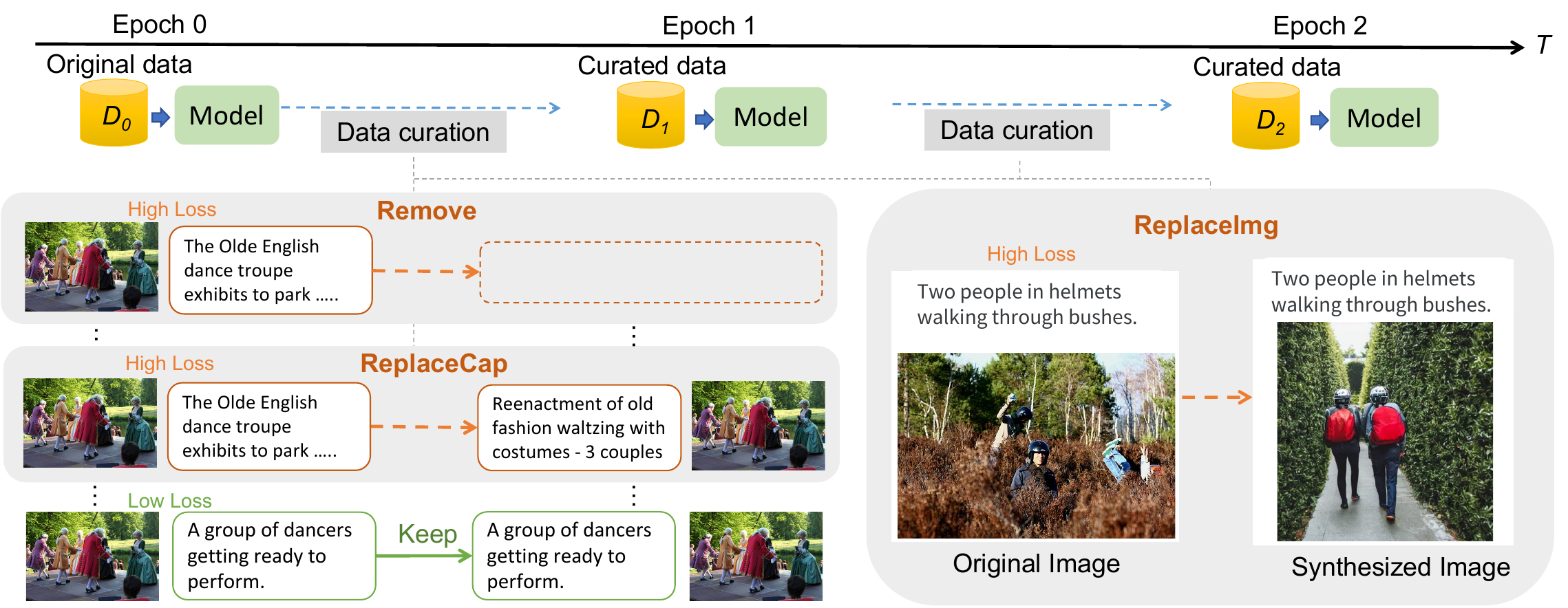}
        \caption{Overview of our data curation methods. For \textsc{Remove}, high loss image-text pairs are removed; for \textsc{ReplaceCap}, the image is paired with an alternative caption from the original dataset; for \textsc{ReplaceImg}, captions of original images are used as prompts for text-to-image generation to synthesize new image--text pairs. We experiment with both options of replacing the image only, or pair another relevant caption to the synthesized image.}
        \label{fig:overview}
        \vspace{-6pt}
    \end{figure*}

    Our main goal is to assess whether actively curating image-caption pairs during training can improve image captioning models. There are many reasons for the existence of difficult samples, including mismatches between the image-caption or inconsistencies between the image and caption~\citep{2020TextAU}, e.g. the caption includes mentions of entities that cannot be seen in the image. For clarity in what follows, let $\mathcal{D}$ be an image captioning training dataset with $K$ images, and let I$_k$ be the $k$-th image. Each image is paired with $J$ captions; let C$^j_k$ be $j$th caption of image $k$, and thus, let (I$_k$, C$_k^j$) be an image--caption sample.

    \subsection{Identifying the difficult samples}

        Inspired by scheduling in curriculum learning \citep{bengio2009curriculum,kumar2010self}, we assume that difficult training samples can be automatically identified throughout the training process.
        We propose to use the captioning model $\mathcal{M}$ that is being trained on dataset $\mathcal{D}$ to automatically identify such samples. We can readily use this model to calculate the loss of each sample in $\mathcal{D}$ at any point in time, such as at the end of each epoch $t$: $\mathcal{L}_{\mathcal{M}}^t$(I$_k$, C$^j_k$) $\forall j,k$. The samples can be be ranked by their respective losses, providing candidates for samples that may benefit from data curation. In particular, the highest loss samples are targets for our data curation methods. We focus on samples with losses that are either two standard deviations from the mean, or the top $X$\% highest loss samples. The data curation performs dynamic updates to the training dataset $\mathcal{D} \rightarrow \mathcal{D}_1 \rightarrow \cdots  \rightarrow \mathcal{D}_T$. In this way, the training dataset is dynamically updated at the end of each epoch according to the model's current captioning capability at time $t$. We empirically observe that without data curation, the high-loss samples remain high-loss during five epochs of training.\footnote{The leftmost plot in Figure~\ref{fig:loss-distr} shows the empirical distribution of losses in the training samples of the Flickr30K.} %

    \subsection{Curation approaches} \label{sec:curation_approaches}
    We investigate three approaches to dynamically curate the high-loss image-caption pairs: \textsc{Removal}, \textsc{ReplaceCap}, and \textsc{ReplaceImg}. Figure \ref{fig:overview} shows an overview of these approaches.\looseness=-1
    \paragraph*{\textsc{Remove}}\label{para:remove}
        The simplest approach to data curation is to remove the high-loss samples, preventing the samples from confusing the model. In \textsc{Remove}, the high-loss samples are completely removed from the remainder of the training process, reducing the total number of image--caption training samples. 
    
    \paragraph*{\textsc{ReplaceCap}}\label{para:dupCap}
    In \textsc{ReplaceCap}, we simply replace the caption in the image--caption sample with a different caption from the original dataset that describes the image, effectively creating a duplicate. With this method, the total number of samples used to train the model remains the same, as well as the total number of the unique images. This creates a control condition for our  experiments. As an alternative, we also experiment with replacing the original caption with one generated by a language model, which we discussed in Section \ref{para:lm}.

    \paragraph*{\textsc{ReplaceImg}}\label{para:replaceImg}
    In \textsc{ReplaceImg}, we perform data curation using a text-to-image generative model. This has the benefit of training the model on the same total number of samples while exposing it to more unique images.
    In a rapid model-in-the-loop step, we use a text-to-image generation model to synthesize images based on the other sentences that describe the image. We integrate this into training as follows: Given an image $I_k$ in the training data and its captions $\{(I_k, C^1_k), \dots, (I_k, C_k^{J})\}$, we synthesize a new image $\hat{I}_k$ without increasing the total number of samples in the original dataset. 
    Specifically, for image $I_k$, we replace an original high-loss sample $(I_k, C^j_k)$ with the synthesized image-text pair $(\hat{I}_k, C^j_k)$. 

    Given a set of captions that describe an image, there are several options for how to prompt the image generation model (Figure~\ref{fig:prompting} in Appendix). 
    We experiment with three options:
        \begin{itemize}[itemsep=1ex]
            \item Single caption: Each caption is used in isolation to generate a new image. \vspace{-3pt}
            \item Sentence-BERT selection: There is a lot of variety in how different captions describe the same image. Instead of using all captions, we can use a representative caption from the set. This is achieved using the Sentence-BERT \citep{reimers-gurevych-2019-sentence} model to find the caption that is closest to the average embedding of all captions.\vspace{-3pt}
            \item Concatenation: All five captions are concatenated as the text prompt for generation.\vspace{-3pt}
        \end{itemize}
        
        For all three approaches mentioned above, we can append an additional string to the prompt as a \textit{styler} to force a specific style in the generated image (+Styler). The styler used here is: "national geographic, high quality photography, Canon EOS R3, Flickr".\footnote{The styler was chosen by inspecting the generated images, with a preference for photographic outputs and against ``artistic'' outputs, such as sketches and computer art.} Some representative examples of images generated using this technique can be seen in Figure~\ref{fig:examples-loss} in the Appendix.

    \begin{table*}[t]
            \centering
            \centering
            \begin{tabular}{rllrrrrrrrrr}
            \toprule   
              & & &  \multicolumn{4}{c}{BLIP} & \multicolumn{4}{c}{BEiT-3} \\
              \cmidrule(rl){4-7} \cmidrule(l){8-11} 
             & Method & Ratio &B & M & C & CS & B & M & C & CS\\ 
            \cmidrule(l){2-2}  \cmidrule(lr){3-3} \cmidrule(rl){4-7} \cmidrule(rl){8-11}

            \multirow{4}{*}{\rotatebox[origin=c]{90}{Flickr30K}} &
            \cellcolor[gray]{.95}Baseline & \cellcolor[gray]{.95}- & \cellcolor[gray]{.95}37.6 & \cellcolor[gray]{.95}27.2 & \cellcolor[gray]{.95}92.8 & \cellcolor[gray]{.95}78.6
            & \cellcolor[gray]{.95}28.9 & \cellcolor[gray]{.95}27.2 & \cellcolor[gray]{.95}79.3 & \cellcolor[gray]{.95}\textbf{80.4} \\
            &  +Remove & 2 std & {38.6} & \textbf{27.4} & \textbf{95.8} & \textbf{79.2}
                      & {31.4} & {27.1} & \textbf{83.7}  & {80.0} \\ 
            &  +ReplaceCap & 1\% & {37.9} & \textbf{27.4} & {94.5} & {78.9} 
                      & {29.6} & \textbf{27.5} & {80.1} & {80.3} \\
            &  +ReplaceImg & 40\% & \textbf{39.0} & {27.3} & {95.7} & {79.1} 
                          & \textbf{32.0} & {26.9} & {82.4} & {79.1} \\
            \cmidrule(l){2-2}  \cmidrule(lr){3-3} \cmidrule(rl){4-7} \cmidrule(rl){8-11}
            \multirow{4}{*}{\rotatebox[origin=c]{90}{COCO}}  &
            \cellcolor[gray]{.95}Baseline & \cellcolor[gray]{.95}- & \cellcolor[gray]{.95}39.9 & \cellcolor[gray]{.95}30.8 & \cellcolor[gray]{.95}132.0 & \cellcolor[gray]{.95}77.3 
                 & \cellcolor[gray]{.95}39.4 & \cellcolor[gray]{.95}31.1 & \cellcolor[gray]{.95}133.7 & \cellcolor[gray]{.95}77.4 \\
            &  +Remove & 1\% & {40.1} & {30.9} & {132.5} & 77.3 
                      & {39.3} & 31.1 & {133.2} & {77.3} \\
            &  +ReplaceCap & 1\% & \textbf{40.2} & {30.9} & {132.7} & 77.3 
                          & 39.4 & {31.0} & {133.6} & {76.5} \\
            &  +ReplaceImg & 10\% & \textbf{40.2} & \textbf{31.0} & \textbf{133.1} & 77.3 
                          & \textbf{39.6} & \textbf{31.1} & \textbf{134.4}  & \textbf{77.5} \\
              
            \bottomrule
            \end{tabular}
            \caption{Results of finetuning with our data curation methods compared to standard finetuning of BLIP and BEiT-3 on the Flickr30K and COCO datasets. 
            We report \textbf{B}LEU, \textbf{M}eteor, \textbf{C}IDEr, and \textbf{C}LIP\textbf{S}core. Best scores are in \textbf{bold}. \label{tab:finetune}}
        \end{table*}

\section{Experimental Setup}
\label{sec:exp}
      
     \subsection{Data \& Metrics} 
     We evaluate our data curation methods during finetuning on the widely used MS COCO \citep{cocodataset} and Flickr30K \citep{flickr30k} datasets. We report results using the metrics of \textbf{B}LEU \citep{papineni2002bleu}, \textbf{M}ETEOR \citep{denkowski2014meteor}, \textbf{C}IDEr \citep{vedantam2015cider}, and \textbf{C}LIP\textbf{S}core \citep{hessel2021clipscore}.
     \vspace{-6pt}
    
    \subsection{Models \& Implementation} 
    
    \paragraph{Image Captioning Models}
    We study the effectiveness of data curation with two state-of-the-art pretrained vision-language models -- \abr{BLIP} \citep{li2022blip} and BEiT-3 \citep{beit3}. 
    
    We note that BLIP has a captioning and filtering (CapFilt) data augmentation process during its pretraining, where both components were finetuned on the COCO dataset. Therefore we use pretrained checkpoint BLIP$_{CapFilt}$ for Flickr30k and BLIP$_{base}$ for COCO in our experiment, removing the effects of the CapFilt process. We finetune BLIP using a total batch size of $128$ for $5$ epochs on $4\times$A100 GPUs. The BEiT-3 base model is finetuned with the default setups: a total batch size of $256$ for $10$ epochs on $8\times$A100 GPUs.

    \paragraph{Curation Ratio} We tune the amount of data to be curated for each method on the validation data of each dataset using the BLIP model. See Section ~\ref{sec:discussion} for more discussion on the trade-off between the amount of data curation and model performance.
    
    \paragraph{\textsc{ReplaceImg} Text-to-image Generation}
    For text-to-image generation in \textsc{ReplaceImg}, we use the open source Stable Diffusion model~\citep{sd}, which can generate images given a textual prompt. We finetune a Stable Diffusion v1.5 model, using the MS COCO~\citep{cocodataset} dataset with a prompt consisting of a concatenation of all 5 captions, for 15,000 steps with a constant learning rate of \num{1e-5} and a batch size of 32. We experiment different versions of the released Stable Diffusion models and various techniques for generating high-quality images for replacement.\footnote{It is also possible to use API-based models but we chose Stable Diffusion because (i) Stable Diffusion can be integrated directly into our training pipeline using the open source code. And (ii) we estimate that it would cost \$4,176 to run a single experiment on the Flickr30K dataset using DALL·E-2 as of Feburary 1st, 2024.} We find that using a finetuned text-to-image model enhances image captioning performance. See Section~\ref{sec:ablation} for further analysis and ablation.

\section{Results}
    \label{sec:results}

    \paragraph*{Data curation improves captioning}
        
        \begin{figure}[t]
            \includegraphics[width=\columnwidth,trim={5mm 2mm 8mm 4mm},clip]{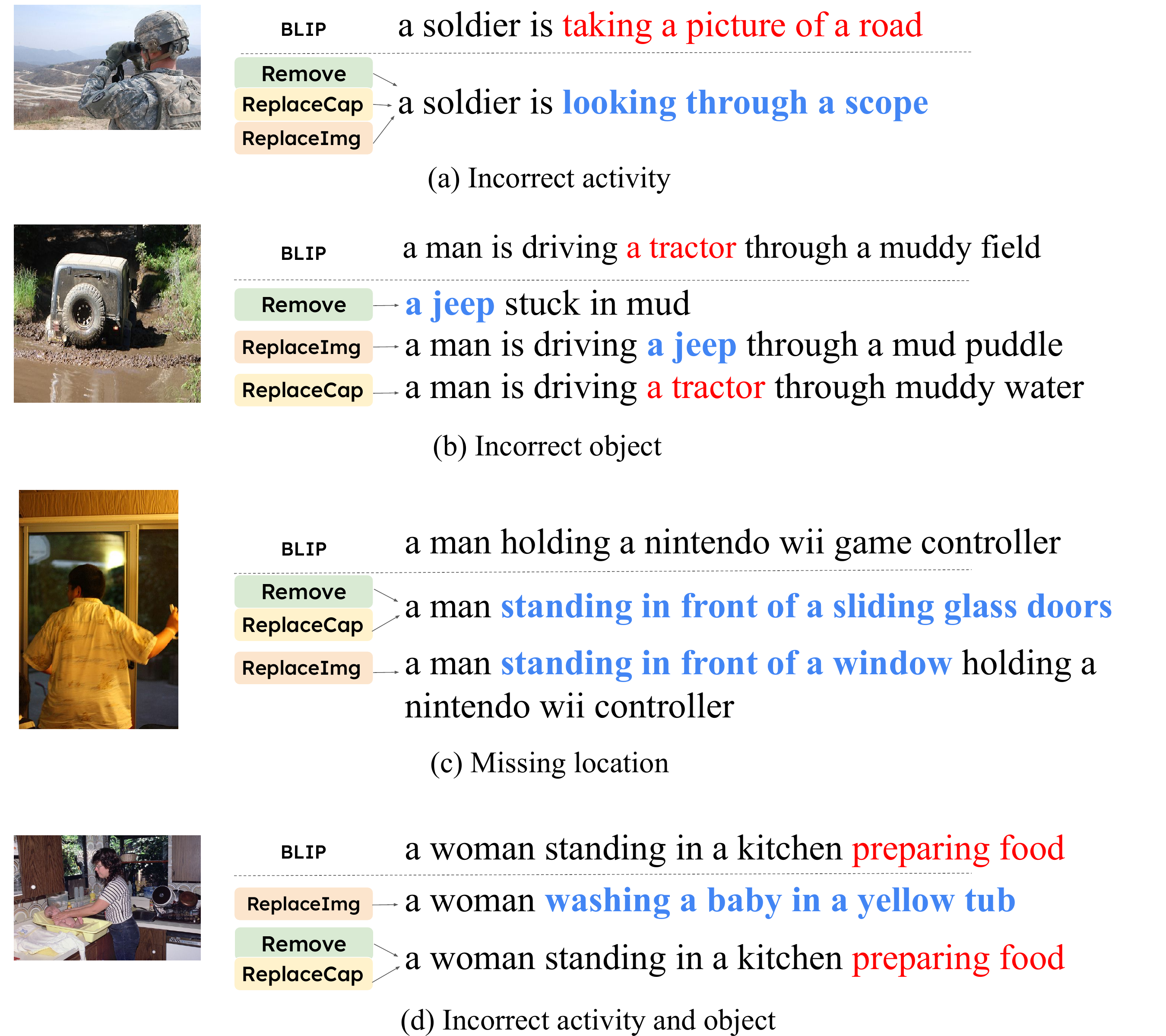}
            \caption{Qualitative examples from the COCO dataset of captions generated by the BLIP model (top), and the same models trained using our data curation methods (bottom). After curation, many of the errors (in red) can be avoided  or fixed (in blue).\label{fig:examples}}
            \vspace{-6pt}
        \end{figure}

        \begin{figure*}[t]
            \centering
            \begin{subfigure}{0.5\textwidth}
            \centering
                \includegraphics[max size={\linewidth}{0.6\textheight}]{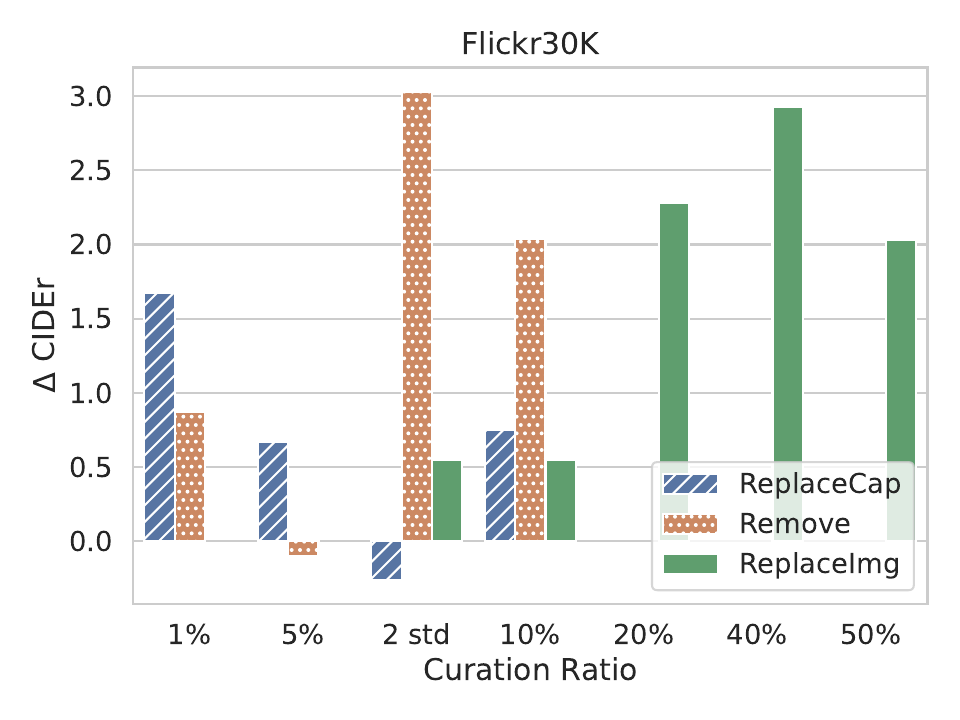}
            \end{subfigure}%
            \begin{subfigure}{0.5\textwidth}
            \centering
                \includegraphics[max size={\linewidth}{0.6\textheight}]{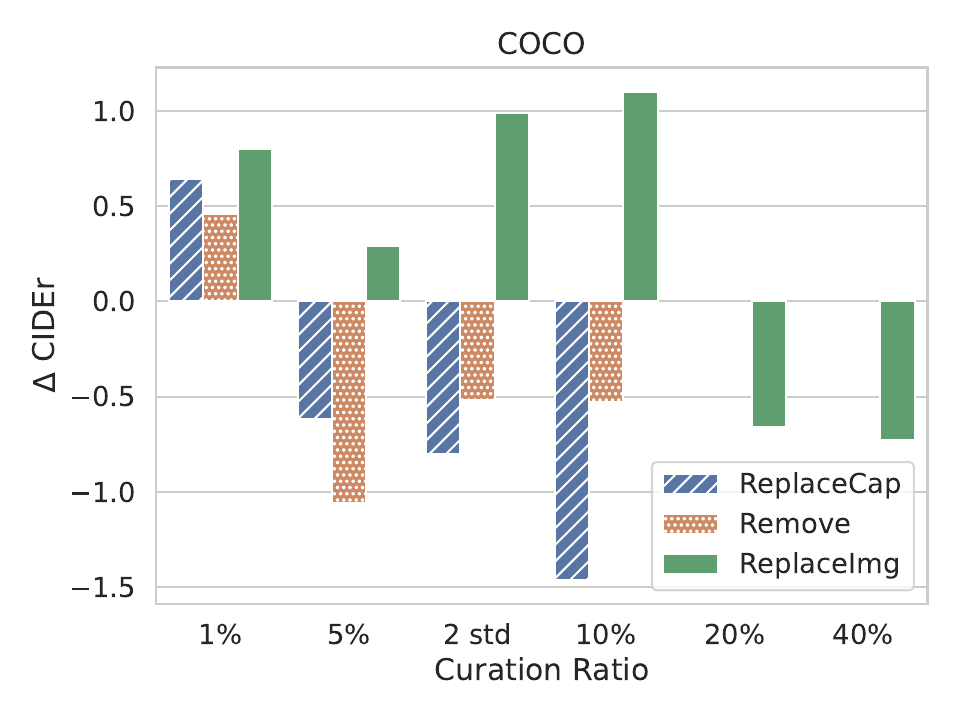}
            \end{subfigure}
            \caption{Effects of varying the amount of data curated. We observe that Flickr30K needs more curation (40\% \textsc{ReplaceImg} or 2 std \textsc{Remove}) than COCO (10\% \textsc{ReplaceImg} or 1\% \textsc{ReplaceCap}). Flickr30K benefits more from removing high-loss training samples, indicating the original dataset may be noisier than MS COCO. For the 2 std approach, the number of samples curated is not fixed after each epoch and varies between 5\% to 10\%. \label{fig:sd-rep-ratio}}
        \end{figure*}

        Table~\ref{tab:finetune} shows the results for the Flickr30K and COCO datasets with the BLIP and BEiT-3 models. 
        The main conclusion is that better model performance can almost always be achieved using data curation. 
        
        For Flickr30K, it can be seen that \textsc{Remove} (2 std) and \textsc{ReplaceImg} (40\%) perform similarly well with a 2.9--3 CIDEr points improvement. The \textsc{ReplaceCap} method only improves performance by 1.7 CIDEr points when applied to the top 1\% of high-loss samples. For COCO, the best performing approach is \textsc{ReplaceImg} with a curation ratio of 10\%, bringing a 1.1 CIDEr point improvement over the baseline. \textsc{ReplaceCap} and \textsc{Remove} both work best when curating the top 1\% of high-loss samples, bringing smaller improvements of 0.5--0.7 CIDEr points. Qualitative examples of the improvements can be seen in Figure~\ref{fig:examples}.
    
    \paragraph*{Generalization to different VL models}
        We also verify that our data curation methods generalize to other models by implementing them in the BEiT-3 model. More specifically, we used exactly the same curation ratio that gained improvements for BLIP. As shown in Table~\ref{tab:finetune}, where \textsc{Remove} is also the most efficient approach for better captioning on Flickr30K, and \textsc{ReplaceImg} improves the most for COCO. This shows that the curation methods can be readily applied to other state-of-the -art vision-language models and the curation ratios are transferable. We note that since BEiT-3 includes COCO in pretraining, the \textsc{Remove} and \textsc{ReplaceCap} methods are not beneficial.

   \section{Discussion}
    \label{sec:discussion}
    \paragraph{Curation amount matters} \label{para:amount}
        
        The amount of data curated is an important hyperparameter. In addition to the best results reported above, we present finer-grained results of varying the amount of data curation. For \textsc{Remove} and \textsc{ReplaceCap}, we explore curating the top 1\%, 5\% and 10\% of high-loss samples. For \textsc{ReplaceImg}, we explore 10\%--80\% curation ratios. In addition to fixed X\% ratios, we also intereven on samples that have losses two standard deviations worse than the mean. 
        
        \begin{figure}[!t]
            \centering
            \vspace{-6pt}
                \includegraphics[max size={0.8\linewidth}{0.7\textheight},trim={5mm 5mm 5mm 3mm}]{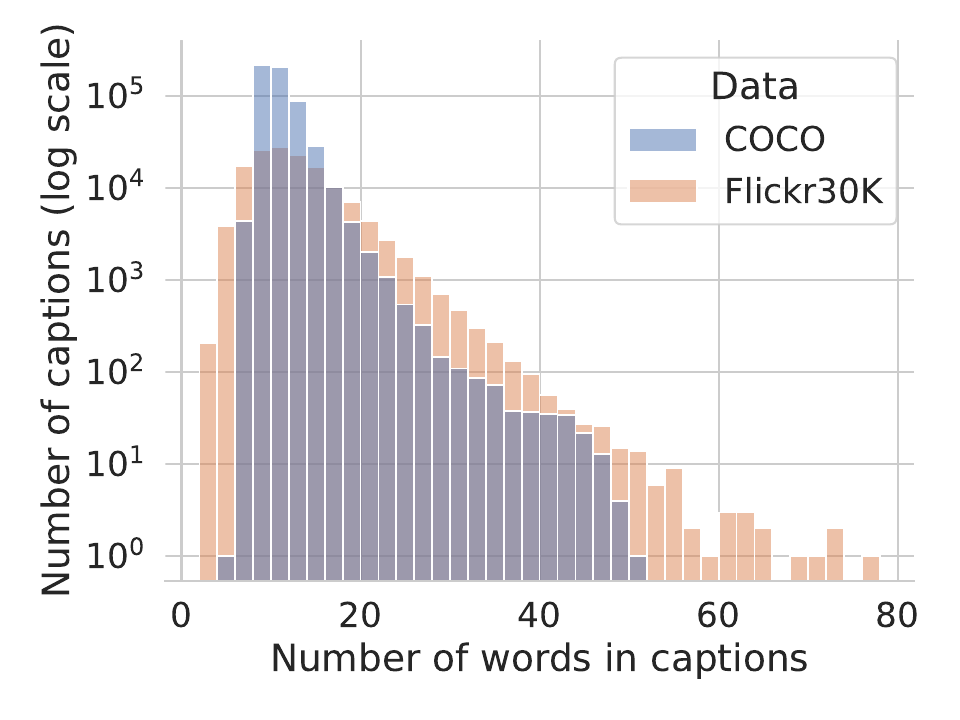}
                \caption{Distribution of caption lengths.\label{fig:captionlen}}
            \vspace{-12pt}
        \end{figure}

        \begin{figure*}[t]
            \centering
            \vspace{-6pt}
                \includegraphics[max size={0.9\linewidth}{0.5\textheight},trim={5mm 1mm 5mm 5mm}]{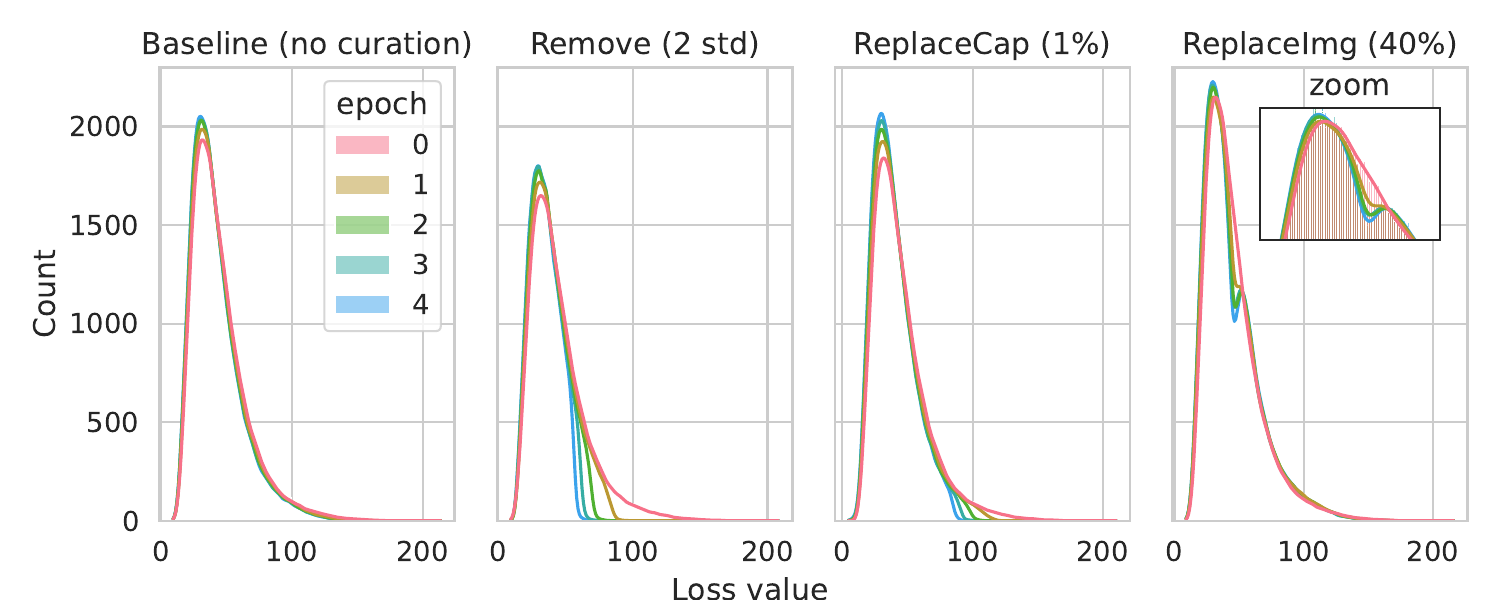}
                \caption{Different curation methods change the loss distribution of training samples over epochs for Flickr30K. In contrast, in the absence of data curation (the leftmost plot), high-loss samples consistently retain their high-loss status throughout the training process.\label{fig:loss-distr}}
            \vspace{-6pt}
        \end{figure*} 

        The results of this analysis are shown in Figure~\ref{fig:sd-rep-ratio}. While the effective curation ratio for different curation approach ranges from 1\%-50\% for Flickr30K, COCO benefits from \textsc{ReplaceImg} on less than 10\% of the top loss samples, and the effective curation ratio for \textsc{Remove} and \textsc{ReplaceCap} stops at 1\%. This indicates that Flickr30K may contain more noisy samples than the MS COCO dataset. Compared to MS COCO, Flickr30K contains more samples with long captions (Figure~\ref{fig:captionlen}), which may include overly-specific details that are inconsistent with other captions and are hard for the model to learn (Figure~\ref{fig:examples-highloss}). Through our curation-based finetuning, these samples can be effectively identified, removed or replaced, which indicates that our method is efficient when training with noisy datasets. We note that curating more than 50\% of the data does not benefit training and actually harms performance.

        \paragraph*{Curation changes training distributions}

        \begin{figure}[t]
            \centering
            \vspace{-6pt}
                \includegraphics[max size={0.95\linewidth}{\textheight}]{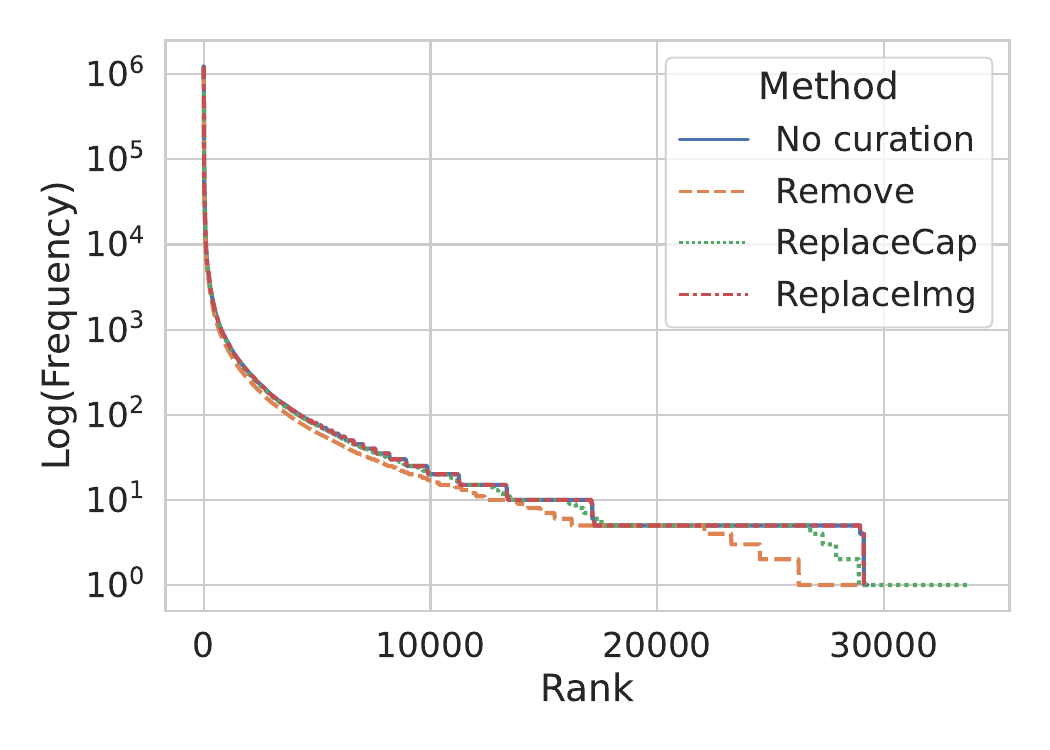}
                \caption{Zipfian distribution of words in Flickr30K training samples for different curation approaches. Note the clear changes made to the tail by \textsc{Remove}. \label{fig:zipf}}
            \vspace{-6pt}
        \end{figure}

        We examine the loss distributions of training samples across epochs for each curation method to understand their impact on the training process (Figure~\ref{fig:loss-distr}). These losses are computed after each epoch using the current model parameters, with high-loss samples being targeted for the subsequent curation step. 
        For the \textsc{Remove} approach, training samples with loss that are two standard deviations worse than the mean are dynamically removed during training, leading to the shrinking tail of the loss distribution. \textsc{ReplaceImg} gradually reduces losses, resulting in the losses forming a mixture of Gaussians consisting of the original image-text pairs and the those with synthesized images. Going beyond just the losses of the training samples, we also inspect the distributions of the words in the training captions for the curation methods. Figure~\ref{fig:zipf} shows these distributions, where it can be seen that \textsc{Remove} reduces low-frequency and singleton words during training, while \textsc{ReplaceCap} increases the counts of some lower-frequency words while removing singletons. By definition, \textsc{ReplaceImg} only changes the distribution of the images used to train the model, and as such, does not change the distribution of the words in the training data.

        \begin{figure}[t]
            \centering
                \includegraphics[max size={0.9\linewidth}{0.7\textheight},trim={10mm 5mm 5mm 5mm},clip]{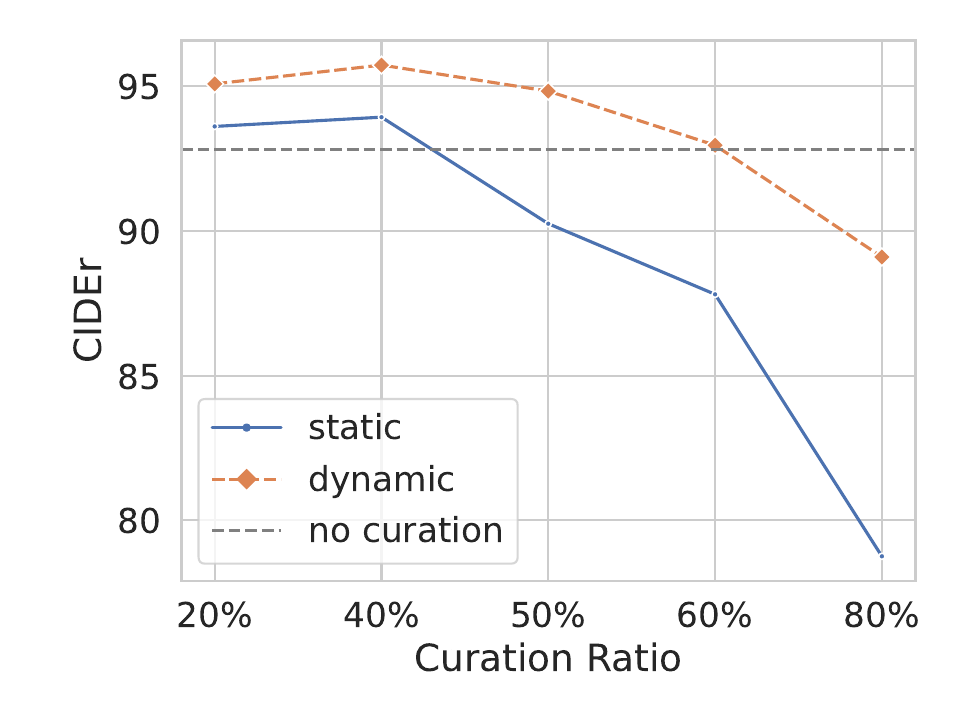}
                \caption{Dynamic versus static replacement for \textsc{ReplaceImg} using BLIP on the Flickr30K dataset, as a function of the number of samples replaced. \label{fig:dynamic-static}}
              \vspace{-6pt}
        \end{figure} 
    
    \paragraph*{The efficacy of dynamic replacement}

    \begin{figure*}[!t]
            \centering
            \begin{minipage}[t]{0.5\textwidth}
                \centering
                \subcaptionbox{Distribution of text-to-image generation errors. \label{fig:sd-errors}}{
                \includegraphics[width=0.91\textwidth]{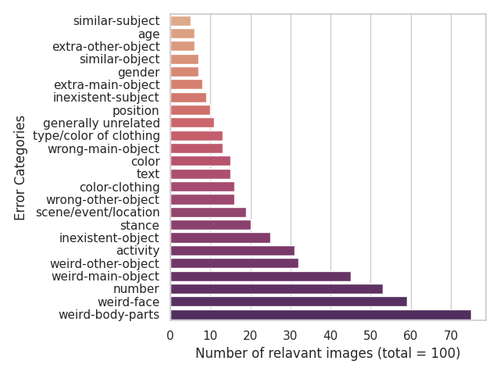}}
            \end{minipage}\hfill
            \begin{minipage}[t]{0.45\textwidth}
                \centering
                \subcaptionbox{Human evaluation versus CLIPScore.\label{fig:clip-correlation}}{
                \includegraphics[width=\textwidth]{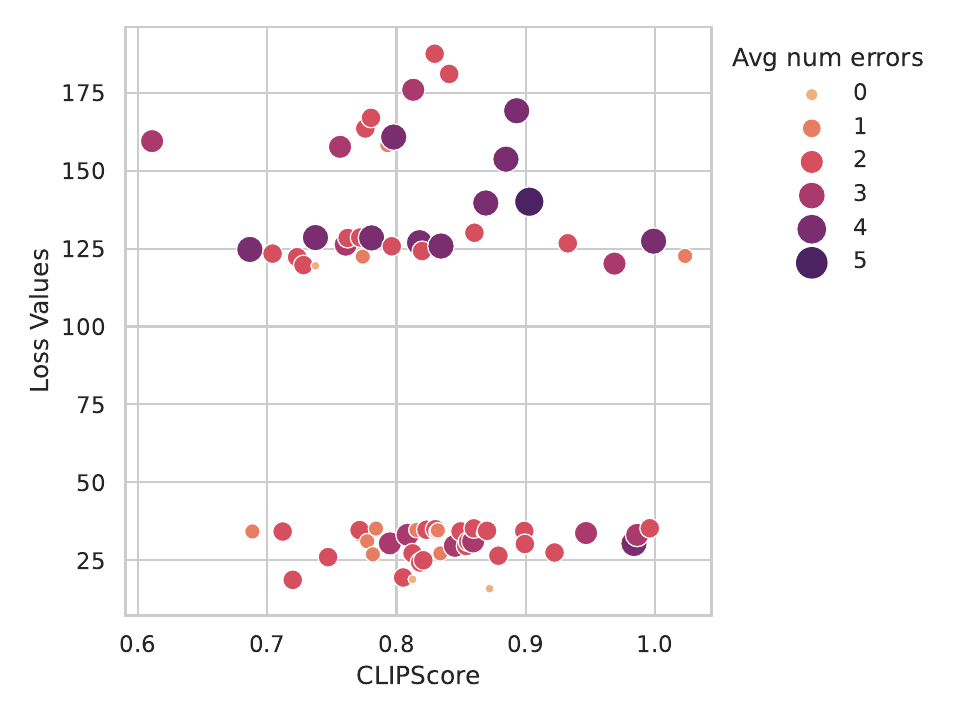}}
                \end{minipage}
            \caption{Results of the human study of the errors made by the Stable Diffusion model in 100 images. The images used in the study were chosen to represent either low or high model loss. (a) Histogram of the number of errors annotated in each category. The most frequently occurring annotations concern weird deformations in the expected objects or humans. (b) Relationship between average number of identified errors by human annotations for each synthesized image and its captioning loss with regard to original captions. More errors are identified in images of higher loss. However, CLIPScore appears to fail in validating qualities of the synthesized images, as the score ranges are almost identical for samples that contain more errors. \label{fig:sd-analysis}}
        \end{figure*}

        Using training loss values as an effective indicator, we dynamically curate on the training samples identified as challenging. In \textsc{ReplaceImg}, another static approach is to replace the identical images, i.e. $I_k$ in $\{(I_k, C^1_k), \dots, (I_k, C_k^{J})\}$, with unique synthesized images before training, instead of updating the training samples while training. With static image replacement, for each of the reference captions, we replace their original image with a generated image. Static replacement with 20\%--80\% curation ratio corresponds to replacing images for one--four captions of the original five. The 50\% replacement ratio mimics a fair coin-flip, where for each of the text-image samples, there is 50\% probability for the image to be replaced by a synthesized image. 

        We compare the efficacy of these two approaches in Figure~\ref{fig:dynamic-static}. When evaluating on the original 1k validation set, we see that for both approaches, incorporating synthesized images of 20\% or 40\% can assist finetuning and achieves higher CIDEr scores. Nevertheless, dynamic image replacement consistently performs better than the static method, showing focusing on the hard samples is effective. For both replacement methods, performance starts to decrease when the curation ratio is too high. This may indicate that when incorporating too many images from the synthetic distribution, the gap increases between the training and evaluation sets.

    \paragraph*{Replacing captions with LM generations}
    \label{para:lm}
    As an alternative to the \textsc{ReplaceCap} method, we investigate the utility of replacing the captions with those generated by a language model (LM). Inspired by the approach in \citet{ramos-etal-2023-lmcap}, we prompt the XGLM-2.9B model~\citep{lin2022few} with few-shot examples to generate a new caption. The LM generated caption is then paired with the image as the curated sample. We evaluate on Flickr30K using both models, applying the same curation ratio of $1\%$ as \textsc{ReplaceCap}. The results presented in Table~\ref{tab:lm-replace} indicate that this approach can serve as a viable alternative to \textsc{ReplaceCap}, consistently outperforming baselines for both models. Please refer to Appendix \ref{app:xglm} for more implementation details.

    \begin{table}[]
        \begin{tabular}{lrrrr}
        \toprule
              & \multicolumn{2}{c}{BLIP} & \multicolumn{2}{c}{BEiT3} \\ 
              \cline{2-5}
                Method & B &  C  & B  & C \\ \hline
        \rowcolor[gray]{.95} 
        Baseline      & 37.6        & 92.8  & 29.8        & 79.3 \\
        +ReplaceCap   & \textbf{37.9} & \textbf{94.5} & 29.6  & 80.1 \\
        +ReplaceLMCap & 37.5 & 93.4   & \textbf{31.2}& \textbf{83.2} \\
        \bottomrule
        \end{tabular}
        \caption{Comparing caption replacement with LM generation to \textsc{ReplaceCap} on Flickr30K. Both methods improve over baseline for BLIP and BEiT-3. \label{tab:lm-replace}}
        \vspace{-8pt}
        \end{table}
    
    \paragraph*{Human Study: Errors made by text-to-image generation models}\label{subsec:humanstudy}
        
        To assess the quality of the generated images and their alignment with human judgments, we perform a human study to evaluate the errors present in the synthesized images. This will serve to better understand any shortcomings with the \textsc{ReplaceImg} curation that is not captured by automatic evaluation measures.
    
        We first ranked synthesized images by model loss from the 1K images in the COCO validation set. 
        We then sampled a subset for human annotation using the top and bottom $50$ images based on their loss using our fine-tuned captioning model. These images are uniformly divided into $5$ sets, each containing $20$ images with equal number of the high loss ones and the low loss ones. The data was annotated by 12 people, members of a university research lab with a basic understanding of text-to-image generation but no knowledge of the bi-modal distribution of images. The annotators were asked to categorize the errors in the synthesized images, given both the image and the reference sentences that were used to generate the images. Each participant annotated one set images.
        
        Starting from the categories defined by~\citet{MiltenburgE17}, we defined 25 error categories including color, number mismatches, and errors related to people and objects in the images. Please see the user interface and more details in the Appendix~\ref{appendix:user-study}. We analyze the human judgements for the images that have at least three annotations, yielding 74 unique images.
        
        As shown in Figure~\ref{fig:sd-errors}, the most common problem of the synthesized images are that they often generate weird face or body parts, which makes the images less natural or pleasant. The text-to-image generation model is also weak at generating the correct number of people or objects. From Figure~\ref{fig:clip-correlation} we confirm the quality of our collected annotations that high loss figures often contain more errors on average. Furthermore, we note that CLIPScore is insensitive to these types of errors, indicating its limited capability of evaluating quality of generated images. Additional examples can be found in Figure~\ref{fig:examples-loss} in the Appendix.

\section{Further Analysis}
\label{sec:ablation}
    With the human study revealing the failure modes of the text-to-image model, we now provide insights on various techniques that are proved useful for improving image relevance in curating the image captioning datasets.

        \paragraph*{Round-trip captioning evaluation}\label{para:roundtrip-caption}

        Most previous work in text-to-image generation uses image-oriented measures like FID \citep{Heusel2017GANsTB} or CLIPScore \citep{hessel2021clipscore}. However, these measures are not suitable for our purpose as they are claimed to lack alignment with perceptual quality~\citep{Saharia2022PhotorealisticTD}. We also found that CLIPScore cannot distinguish between low- and high-loss samples in captioning (Figure~\ref{fig:sd-analysis}). 

\begin{figure}[!tb]
            \centering
            \includegraphics[width=0.45\textwidth,trim={1mm 0mm 2mm 2mm},clip]{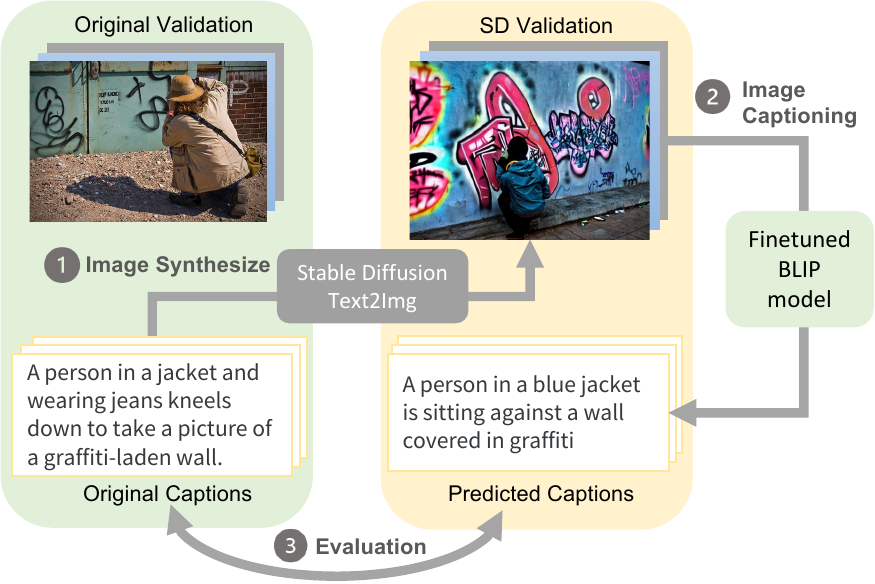}
            \captionof{figure}{Round-trip captioning evaluation. \label{fig:roundtrip-caption}}
        \end{figure}

        Alternatively, similar to \citet{hong2018inferring}, we use a fixed model to generate captions for synthesized images and then compare them to original captions in a three-step process (Figure~\ref{fig:roundtrip-caption}): (1) Generating images from validation set captions; (2) Predicting captions for the generated images using a strong image-captioning model; here we use BLIP fine-tuned on the COCO dataset but any other strong captioning model could be used instead. (3) Comparing the predicted captions with the original captions. The assumption is that if the generated images are of similar quality to the originals, the resulting captions will also be similar. 
        
        \paragraph*{Ablation on text-to-image variants} Evaluating with round-trip captioning, we conduct an ablation study on variants of text-to-image generation models. Table~\ref{tab:roundtrip} summarizes the evaluation results on the Flickr30K dataset. Specifically, we experiment with different versions of the Stable Diffusion models; prompt the diffusion models with various approaches (Section \ref{sec:curation_approaches}); and compare the generation performance between the finetuned text-to-image model and the pretrained ones. The results show that Stable Diffusion v1.5 finetuned on COCO outperforms the other variants, when prompted with the concatenation of all five captions, with the addition of the styler. For the details of the model variants, please refer to Appendix~\ref{appendix:prompt}.

        \begin{table}
        \resizebox{\linewidth}{!}{
         \begin{tabular}{lllccc}%
            \toprule   
            Model & FT & Prompt & B & C & M \\
            \midrule
            \rowcolor[gray]{.95}
            \multicolumn{3}{l}{Upper-bound} & 37.6 & 27.2 & 57.1 \\
            SD 1.5 & - & concat & 31.0 & 24.7 & 52.5 \\
            SD 1.5 & - & \hspace{1ex}+ styler & 30.8 & 24.2 & 52.5 \\
            SD 1.5 &  F & \hspace{1ex}+ styler  & \textbf{33.5} & \textbf{25.0} & \textbf{53.5} \\
            SD 1.5 &  F & SBERT + styler & 30.6 & 24.1 & 52.0 \\
            SD 2.0 & - & concat + styler & 31.2 & 24.8 & 52.0 \\
            \bottomrule
            \end{tabular}}
        \caption{Round-trip captioning evaluation on Flickr30K with different Stable Diffusion models, prompts, and fine-tuning. F indicates that the model is finetuned. We report \textbf{B}LEU, \textbf{C}IDEr, \textbf{M}eteor. \label{tab:roundtrip}}
        \end{table}

\section{Conclusion}
\label{sec:conclusion}

In this paper, we have shown a simple, yet effective, data curation framework that can improve the performance of image captioning models. We investigated three approaches to data curation that dynamically update the training dataset based on high-loss image-caption samples. The methods involved either removing a sample, replacing the caption in a sample, or generating a new image from existing captions. Experimental results on the Flickr30K and MS COCO datasets show the effectiveness of these approaches to data curation without increasing the total size of the training dataset. A deeper analysis of the images synthesized by the text-to-image model shows frequent errors on generating objects of a certain amount or color, and struggles with human body features. A human evaluation of the errors in those images shows a clear difference in images with high or low losses.

In the future, we expect that better text-to-image generation models will lead to further improvements from using synthesized images to train image captioning models. From our insights in Appendix~\ref{para:combine}, there is also significant promise on building a hybrid model combining different curation methods. We believe that a more sophisticated learning scheme leveraging multiple methods will offer more flexibility when curating the dataset. We plan on verifying whether these findings extend to other image captioning models. Moreover, we are also interested in applying the same framework to other multimodal tasks, especially those with under-complete datasets that cannot comprehensively cover the distributional space due to the cost of crowd-sourcing enough data, e.g. visual question answering, or visually-grounded dialog. 

\section*{Limitations}
 As \citet{nguyen2023improving} has successfully improved the quality of the pretraining dataset by using an state-of-the-art BLIP-2 model to generate better captions, we would expect that our curation strategies to be scaled and adapted also to vision-language pretraining, which however is limited by research resources and therefore not explored in the scope of this paper. Currently our data curation methods also rely on state-of-the art pretrained models for both image understanding and text-to-image generation.
 
In our study, we explore how the application of various curation approaches impacts the downstream image captioning performance under different curation ratios. While we predefine the curation ratio for our experiments in this paper, it is desirable for curation methods to be more readily applicable if the curation ratio can be automatically determined. 

Moreover, while we take an online approach to data curation, our current approach is upper bounded in speed and performance of the text-to-image generation model. This might be a large bottle neck for adapting the strategy for more complicated vision-and-language tasks.

\section*{Ethics Statement}

Text-to-image generation is controversial in the broader AI and ethics community\citep{carlini2023extracting}. For example, it can generate images according to gender or racial stereotypes, which may prove harmful to members of those communities~\citep{LI2022107286}. While have not yet been observed in the vision-language domain, \citet{shumailov2023curse} provide evidence that the use of synthetic data from generative models like large language models can introduce a potential risk of data quality degradation.

In this paper, we use text-to-image to improve the quality of an image captioning model, given a specific set of crowd-sourced captions. Those captions may themselves contain harmful stereotypes that would become more prevalent in our dynamically updated training datasets. 
As we dynamically update the model with new images based on loss values, we remove the water-marker in our generated images to prevent information leak to the model. Use of the synthesized images will strictly follow community guidelines. 

While developing our curation methods that involve text-to-image generation for image replacement, we employed the stable-diffusion v1.5 model~\cite{sd}, which was trained on the LAION-5B dataset. We note that we were unaware of any investigation into illegal material in the dataset~\citep{Thiel_2023}. Hence, we emphasize that our proposed framework is compatible with any other text-to-image models trained on more reliable datasets. Taking this in to consideration, we encourage researchers to explore and apply alternative text-to-image models when incorporating the curation techniques in their future work.

\section*{Acknowledgement}
 We thank Jiaang Li, Lei Li, and the CoAStal and LAMP groups for feedback. Wenyan Li is supported by the Lundbeck Foundation (BrainDrugs grant: R279-2018-1145) and by Innovation Fund Denmark in the context of AI4Xray project. Jonas F. Lotz is funded by the ROCKWOOL Foundation (grant 1242). This work was supported by a research grant (VIL53122) from VILLUM FONDEN.

\bibliography{anthology,custom}

\begin{thebibliography}{48}
\expandafter\ifx\csname natexlab\endcsname\relax\def\natexlab#1{#1}\fi

\bibitem[{Alsharid et~al.(2021)Alsharid, El-Bouri, Sharma, Drukker, Papageorghiou, and Noble}]{Alsharid21}
Mohammad Alsharid, Rasheed El-Bouri, Harshita Sharma, Lior Drukker, Aris~T. Papageorghiou, and J.~Alison Noble. 2021.
\newblock \href {https://doi.org/10.1109/ISBI48211.2021.9434055} {A course-focused dual curriculum for image captioning}.
\newblock In \emph{2021 IEEE 18th International Symposium on Biomedical Imaging (ISBI)}.

\bibitem[{Atliha and {\v S}e{\v s}ok(2020)}]{2020TextAU}
Viktar Atliha and Dmitrij {\v S}e{\v s}ok. 2020.
\newblock Text augmentation using bert for image captioning.
\newblock \emph{Applied Sciences}.

\bibitem[{Ayyubi et~al.(2023)Ayyubi, Lokesh, Zareian, Wu, and Chang}]{ayyubi-etal-2023-learning}
Hammad Ayyubi, Rahul Lokesh, Alireza Zareian, Bo~Wu, and Shih-Fu Chang. 2023.
\newblock \href {https://doi.org/10.18653/v1/2023.findings-acl.846} {Learning from children: Improving image-caption pretraining via curriculum}.
\newblock In \emph{Findings of the Association for Computational Linguistics: ACL 2023}.

\bibitem[{Azizi et~al.(2023)Azizi, Kornblith, Saharia, Norouzi, and Fleet}]{azizi2023synthetic}
Shekoofeh Azizi, Simon Kornblith, Chitwan Saharia, Mohammad Norouzi, and David~J. Fleet. 2023.
\newblock \href {http://arxiv.org/abs/2304.08466} {Synthetic data from diffusion models improves imagenet classification}.

\bibitem[{Baltru{\v{s}}aitis et~al.(2018)Baltru{\v{s}}aitis, Ahuja, and Morency}]{baltruvsaitis2018multimodal}
Tadas Baltru{\v{s}}aitis, Chaitanya Ahuja, and Louis-Philippe Morency. 2018.
\newblock Multimodal machine learning: A survey and taxonomy.
\newblock \emph{IEEE transactions on pattern analysis and machine intelligence}, 41(2):423--443.

\bibitem[{Bengio et~al.(2009)Bengio, Louradour, Collobert, and Weston}]{bengio2009curriculum}
Yoshua Bengio, J{\'e}r{\^o}me Louradour, Ronan Collobert, and Jason Weston. 2009.
\newblock Curriculum learning.
\newblock In \emph{Proceedings of the 26th annual international conference on machine learning}, pages 41--48.

\bibitem[{Bernardi et~al.(2016)Bernardi, Cakici, Elliott, Erdem, Erdem, Ikizler-Cinbis, Keller, Muscat, and Plank}]{bernardi2016automatic}
Raffaella Bernardi, Ruket Cakici, Desmond Elliott, Aykut Erdem, Erkut Erdem, Nazli Ikizler-Cinbis, Frank Keller, Adrian Muscat, and Barbara Plank. 2016.
\newblock Automatic description generation from images: A survey of models, datasets, and evaluation measures.
\newblock \emph{Journal of Artificial Intelligence Research}, 55:409--442.

\bibitem[{Caffagni et~al.(2023)Caffagni, Barraco, Cornia, Baraldi, and Cucchiara}]{caffagni2023synthcap}
Davide Caffagni, Manuele Barraco, Marcella Cornia, Lorenzo Baraldi, and Rita Cucchiara. 2023.
\newblock Synthcap: Augmenting transformers with synthetic data for image captioning.
\newblock In \emph{International Conference on Image Analysis and Processing}, pages 112--123. Springer.

\bibitem[{Carlini et~al.(2023)Carlini, Hayes, Nasr, Jagielski, Sehwag, Tram{\`e}r, Balle, Ippolito, and Wallace}]{carlini2023extracting}
Nicholas Carlini, Jamie Hayes, Milad Nasr, Matthew Jagielski, Vikash Sehwag, Florian Tram{\`e}r, Borja Balle, Daphne Ippolito, and Eric Wallace. 2023.
\newblock Extracting training data from diffusion models.
\newblock \emph{arXiv preprint arXiv:2301.13188}.

\bibitem[{Chang and Jia(2023)}]{chang2023data}
Ting-Yun Chang and Robin Jia. 2023.
\newblock Data curation alone can stabilize in-context learning.
\newblock In \emph{Proceedings of the 61st Annual Meeting of the Association for Computational Linguistics (Volume 1: Long Papers)}, pages 8123--8144.

\bibitem[{Chen et~al.(2023)Chen, Cao, and Madden}]{chen2023lingua}
Zui Chen, Lei Cao, and Sam Madden. 2023.
\newblock Lingua manga: A generic large language model centric system for data curation.
\newblock \emph{arXiv preprint arXiv:2306.11702}.

\bibitem[{Denkowski and Lavie(2014)}]{denkowski2014meteor}
Michael Denkowski and Alon Lavie. 2014.
\newblock Meteor universal: Language specific translation evaluation for any target language.
\newblock In \emph{Proceedings of the ninth workshop on statistical machine translation}, pages 376--380.

\bibitem[{Dong et~al.(2021)Dong, Long, Xu, and Xiao}]{dong2021CL}
Xinzhi Dong, Chengjiang Long, Wenju Xu, and Chunxia Xiao. 2021.
\newblock \href {https://doi.org/10.1145/3474085.3475439} {Dual graph convolutional networks with transformer and curriculum learning for image captioning}.
\newblock In \emph{Proceedings of the 29th ACM International Conference on Multimedia}. Association for Computing Machinery.

\bibitem[{Hessel et~al.(2021)Hessel, Holtzman, Forbes, Bras, and Choi}]{hessel2021clipscore}
Jack Hessel, Ari Holtzman, Maxwell Forbes, Ronan~Le Bras, and Yejin Choi. 2021.
\newblock {CLIPScore:} a reference-free evaluation metric for image captioning.
\newblock In \emph{EMNLP}.

\bibitem[{Heusel et~al.(2017)Heusel, Ramsauer, Unterthiner, Nessler, and Hochreiter}]{Heusel2017GANsTB}
Martin Heusel, Hubert Ramsauer, Thomas Unterthiner, Bernhard Nessler, and Sepp Hochreiter. 2017.
\newblock Gans trained by a two time-scale update rule converge to a local nash equilibrium.
\newblock In \emph{NIPS}.

\bibitem[{Hong et~al.(2018)Hong, Yang, Choi, and Lee}]{hong2018inferring}
Seunghoon Hong, Dingdong Yang, Jongwook Choi, and Honglak Lee. 2018.
\newblock Inferring semantic layout for hierarchical text-to-image synthesis.
\newblock In \emph{Proceedings of the IEEE conference on computer vision and pattern recognition}, pages 7986--7994.

\bibitem[{Jain et~al.(2023)Jain, Lawrence, Moitra, and Madry}]{jain2023distilling}
Saachi Jain, Hannah Lawrence, Ankur Moitra, and Aleksander Madry. 2023.
\newblock \href {https://openreview.net/forum?id=99RpBVpLiX} {Distilling model failures as directions in latent space}.
\newblock In \emph{The Eleventh International Conference on Learning Representations}.

\bibitem[{Kandpal et~al.(2022)Kandpal, Wallace, and Raffel}]{pmlr-v162-kandpal22a}
Nikhil Kandpal, Eric Wallace, and Colin Raffel. 2022.
\newblock \href {https://proceedings.mlr.press/v162/kandpal22a.html} {Deduplicating training data mitigates privacy risks in language models}.
\newblock In \emph{Proceedings of the 39th International Conference on Machine Learning}, Proceedings of Machine Learning Research. PMLR.

\bibitem[{Kumar et~al.(2010)Kumar, Packer, and Koller}]{kumar2010self}
M~Kumar, Benjamin Packer, and Daphne Koller. 2010.
\newblock Self-paced learning for latent variable models.
\newblock \emph{Advances in neural information processing systems}, 23.

\bibitem[{Lee et~al.(2022)Lee, Ippolito, Nystrom, Zhang, Eck, Callison-Burch, and Carlini}]{lee-etal-2022-deduplicating}
Katherine Lee, Daphne Ippolito, Andrew Nystrom, Chiyuan Zhang, Douglas Eck, Chris Callison-Burch, and Nicholas Carlini. 2022.
\newblock \href {https://doi.org/10.18653/v1/2022.acl-long.577} {Deduplicating training data makes language models better}.
\newblock In \emph{Proceedings of the 60th Annual Meeting of the Association for Computational Linguistics (Volume 1: Long Papers)}, pages 8424--8445, Dublin, Ireland. Association for Computational Linguistics.

\bibitem[{Li et~al.(2022{\natexlab{a}})Li, Li, Xiong, and Hoi}]{li2022blip}
Junnan Li, Dongxu Li, Caiming Xiong, and Steven Hoi. 2022{\natexlab{a}}.
\newblock Blip: Bootstrapping language-image pre-training for unified vision-language understanding and generation.
\newblock In \emph{ICML}.

\bibitem[{Li et~al.(2022{\natexlab{b}})Li, Wan, and Gao}]{LI2022107286}
Minghui Li, Yan Wan, and Jinping Gao. 2022{\natexlab{b}}.
\newblock \href {https://doi.org/https://doi.org/10.1016/j.chb.2022.107286} {What drives the ethical acceptance of deep synthesis applications? a fuzzy set qualitative comparative analysis}.
\newblock \emph{Computers in Human Behavior}, 133:107286.

\bibitem[{Lin et~al.(2014)Lin, Maire, Belongie, Bourdev, Girshick, Hays, Perona, Ramanan, Doll{'{a} }r, and Zitnick}]{cocodataset}
Tsung{-}Yi Lin, Michael Maire, Serge~J. Belongie, Lubomir~D. Bourdev, Ross~B. Girshick, James Hays, Pietro Perona, Deva Ramanan, Piotr Doll{'{a} }r, and C.~Lawrence Zitnick. 2014.
\newblock Microsoft {COCO:} common objects in context.
\newblock \emph{CoRR}, abs/1405.0312.

\bibitem[{Lin et~al.(2022)Lin, Mihaylov, Artetxe, Wang, Chen, Simig, Ott, Goyal, Bhosale, Du et~al.}]{lin2022few}
Xi~Victoria Lin, Todor Mihaylov, Mikel Artetxe, Tianlu Wang, Shuohui Chen, Daniel Simig, Myle Ott, Naman Goyal, Shruti Bhosale, Jingfei Du, et~al. 2022.
\newblock Few-shot learning with multilingual generative language models.
\newblock In \emph{Proceedings of the 2022 Conference on Empirical Methods in Natural Language Processing}.

\bibitem[{Liu et~al.(2021)Liu, Ge, and Wu}]{liu-etal-2021-competence}
Fenglin Liu, Shen Ge, and Xian Wu. 2021.
\newblock \href {https://doi.org/10.18653/v1/2021.acl-long.234} {Competence-based multimodal curriculum learning for medical report generation}.
\newblock In \emph{Proceedings of the 59th Annual Meeting of the Association for Computational Linguistics and the 11th International Joint Conference on Natural Language Processing (Volume 1: Long Papers)}, pages 3001--3012, Online. Association for Computational Linguistics.

\bibitem[{Nguyen et~al.(2023)Nguyen, Gadre, Ilharco, Oh, and Schmidt}]{nguyen2023improving}
Thao Nguyen, Samir~Yitzhak Gadre, Gabriel Ilharco, Sewoong Oh, and Ludwig Schmidt. 2023.
\newblock Improving multimodal datasets with image captioning.
\newblock \emph{arXiv preprint arXiv:2307.10350}.

\bibitem[{Nichol et~al.(2022)Nichol, Dhariwal, Ramesh, Shyam, Mishkin, McGrew, Sutskever, and Chen}]{nichol2022glide}
Alex Nichol, Prafulla Dhariwal, Aditya Ramesh, Pranav Shyam, Pamela Mishkin, Bob McGrew, Ilya Sutskever, and Mark Chen. 2022.
\newblock \href {http://arxiv.org/abs/2112.10741} {Glide: Towards photorealistic image generation and editing with text-guided diffusion models}.

\bibitem[{Nichol and Dhariwal(2021)}]{nichol2021icml}
Alexander~Quinn Nichol and Prafulla Dhariwal. 2021.
\newblock Improved denoising diffusion probabilistic models.
\newblock In \emph{ICML}, volume 139, pages 8162--8171. {PMLR}.

\bibitem[{Papineni et~al.(2002)Papineni, Roukos, Ward, and Zhu}]{papineni2002bleu}
Kishore Papineni, Salim Roukos, Todd Ward, and Wei-Jing Zhu. 2002.
\newblock Bleu: a method for automatic evaluation of machine translation.
\newblock In \emph{Proceedings of the 40th annual meeting of the Association for Computational Linguistics}, pages 311--318.

\bibitem[{Ramesh et~al.(2022)Ramesh, Dhariwal, Nichol, Chu, and Chen}]{dalle2}
Aditya Ramesh, Prafulla Dhariwal, Alex Nichol, Casey Chu, and Mark Chen. 2022.
\newblock \href {http://arxiv.org/abs/2204.06125} {Hierarchical text-conditional image generation with clip latents}.

\bibitem[{Ramos et~al.(2023)Ramos, Martins, and Elliott}]{ramos-etal-2023-lmcap}
Rita Ramos, Bruno Martins, and Desmond Elliott. 2023.
\newblock \href {https://aclanthology.org/2023.findings-acl.104} {{LMC}ap: Few-shot multilingual image captioning by retrieval augmented language model prompting}.
\newblock In \emph{Findings of the Association for Computational Linguistics: ACL 2023}. Association for Computational Linguistics.

\bibitem[{Reimers and Gurevych(2019)}]{reimers-gurevych-2019-sentence}
Nils Reimers and Iryna Gurevych. 2019.
\newblock \href {https://doi.org/10.18653/v1/D19-1410} {Sentence-{BERT}: Sentence embeddings using {S}iamese {BERT}-networks}.
\newblock In \emph{Proceedings of the 2019 Conference on Empirical Methods in Natural Language Processing and the 9th International Joint Conference on Natural Language Processing (EMNLP-IJCNLP)}, pages 3982--3992, Hong Kong, China. Association for Computational Linguistics.

\bibitem[{Rogers(2021)}]{Rogers2021ChangingTW}
Anna Rogers. 2021.
\newblock \href {https://api.semanticscholar.org/CorpusID:235248305} {Changing the world by changing the data}.
\newblock In \emph{Annual Meeting of the Association for Computational Linguistics}.

\bibitem[{Rombach et~al.(2022)Rombach, Blattmann, Lorenz, Esser, and Ommer}]{sd}
Robin Rombach, A.~Blattmann, Dominik Lorenz, Patrick Esser, and Bj{\"o}rn Ommer. 2022.
\newblock High-resolution image synthesis with latent diffusion models.
\newblock \emph{2022 IEEE/CVF Conference on Computer Vision and Pattern Recognition (CVPR)}, pages 10674--10685.

\bibitem[{Saharia et~al.(2022{\natexlab{a}})Saharia, Chan, Saxena, Li, Whang, Denton, Ghasemipour, Ayan, Mahdavi, Lopes, Salimans, Ho, Fleet, and Norouzi}]{saharia2022photorealistic}
Chitwan Saharia, William Chan, Saurabh Saxena, Lala Li, Jay Whang, Emily Denton, Seyed Kamyar~Seyed Ghasemipour, Burcu~Karagol Ayan, S.~Sara Mahdavi, Rapha~Gontijo Lopes, Tim Salimans, Jonathan Ho, David~J Fleet, and Mohammad Norouzi. 2022{\natexlab{a}}.
\newblock \href {http://arxiv.org/abs/2205.11487} {Photorealistic text-to-image diffusion models with deep language understanding}.

\bibitem[{Saharia et~al.(2022{\natexlab{b}})Saharia, Chan, Saxena, Li, Whang, Denton, Ghasemipour, Ayan, Mahdavi, Lopes, Salimans, Ho, Fleet, and Norouzi}]{Saharia2022PhotorealisticTD}
Chitwan Saharia, William Chan, Saurabh Saxena, Lala Li, Jay Whang, Emily~L. Denton, Seyed Kamyar~Seyed Ghasemipour, Burcu~Karagol Ayan, Seyedeh~Sara Mahdavi, Raphael~Gontijo Lopes, Tim Salimans, Jonathan Ho, David~J. Fleet, and Mohammad Norouzi. 2022{\natexlab{b}}.
\newblock \href {https://doi.org/10.48550/arXiv.2205.11487} {Photorealistic text-to-image diffusion models with deep language understanding}.
\newblock \emph{arXiv preprint}.

\bibitem[{Schuhmann et~al.(2021)Schuhmann, Vencu, Beaumont, Kaczmarczyk, Mullis, Katta, Coombes, Jitsev, and Komatsuzaki}]{Schuhmann2021LAION400MOD}
Christoph Schuhmann, Richard Vencu, Romain Beaumont, Robert Kaczmarczyk, Clayton Mullis, Aarush Katta, Theo Coombes, Jenia Jitsev, and Aran Komatsuzaki. 2021.
\newblock \href {https://doi.org/10.48550/arXiv.2111.02114} {Laion-400m: Open dataset of clip-filtered 400 million image-text pairs}.
\newblock \emph{arXiv preprint}.

\bibitem[{Sharma et~al.(2018)Sharma, Ding, Goodman, and Soricut}]{sharma-etal-2018-conceptual}
Piyush Sharma, Nan Ding, Sebastian Goodman, and Radu Soricut. 2018.
\newblock \href {https://doi.org/10.18653/v1/P18-1238} {Conceptual captions: A cleaned, hypernymed, image alt-text dataset for automatic image captioning}.
\newblock In \emph{Proceedings of the 56th Annual Meeting of the Association for Computational Linguistics (Volume 1: Long Papers)}, pages 2556--2565, Melbourne, Australia. Association for Computational Linguistics.

\bibitem[{Shumailov et~al.(2023)Shumailov, Shumaylov, Zhao, Gal, Papernot, and Anderson}]{shumailov2023curse}
Ilia Shumailov, Zakhar Shumaylov, Yiren Zhao, Yarin Gal, Nicolas Papernot, and Ross Anderson. 2023.
\newblock The curse of recursion: Training on generated data makes models forget.
\newblock \emph{arXiv preprint arxiv:2305.17493}.

\bibitem[{Song et~al.(2021)Song, Sohl{-}Dickstein, Kingma, Kumar, Ermon, and Poole}]{song2021iclr}
Yang Song, Jascha Sohl{-}Dickstein, Diederik~P. Kingma, Abhishek Kumar, Stefano Ermon, and Ben Poole. 2021.
\newblock Score-based generative modeling through stochastic differential equations.
\newblock In \emph{ICLR}.

\bibitem[{Thiel(2023)}]{Thiel_2023}
David Thiel. 2023.
\newblock \href {https://cyber.fsi.stanford.edu/news/investigation-finds-ai-image-generation-models-trained-child-abuse} {Investigation finds ai image generation models trained on child abuse}.

\bibitem[{van Miltenburg and Elliott(2017)}]{MiltenburgE17}
Emiel van Miltenburg and Desmond Elliott. 2017.
\newblock Room for improvement in automatic image description: an error analysis.
\newblock \emph{CoRR}, abs/1704.04198.

\bibitem[{Vedantam et~al.(2015)Vedantam, Lawrence~Zitnick, and Parikh}]{vedantam2015cider}
Ramakrishna Vedantam, C~Lawrence~Zitnick, and Devi Parikh. 2015.
\newblock Cider: Consensus-based image description evaluation.
\newblock In \emph{Proceedings of the IEEE conference on computer vision and pattern recognition}, pages 4566--4575.

\bibitem[{Wang et~al.(2023)Wang, Bao, Dong, Bjorck, Peng, Liu, Aggarwal, Mohammed, Singhal, Som, and Wei}]{beit3}
Wenhui Wang, Hangbo Bao, Li~Dong, Johan Bjorck, Zhiliang Peng, Qiang Liu, Kriti Aggarwal, Owais~Khan Mohammed, Saksham Singhal, Subhojit Som, and Furu Wei. 2023.
\newblock Image as a foreign language: {BEiT} pretraining for vision and vision-language tasks.
\newblock In \emph{Proceedings of the IEEE/CVF Conference on Computer Vision and Pattern Recognition}.

\bibitem[{Wolf et~al.(2019)Wolf, Debut, Sanh, Chaumond, Delangue, Moi, Cistac, Rault, Louf, Funtowicz et~al.}]{wolf2019huggingface}
Thomas Wolf, Lysandre Debut, Victor Sanh, Julien Chaumond, Clement Delangue, Anthony Moi, Pierric Cistac, Tim Rault, R{\'e}mi Louf, Morgan Funtowicz, et~al. 2019.
\newblock Huggingface's transformers: State-of-the-art natural language processing.
\newblock \emph{arXiv preprint arXiv:1910.03771}.

\bibitem[{Xiao et~al.(2023)Xiao, Xu, and Zhang}]{xiao2023multimodal}
Changrong Xiao, Sean~Xin Xu, and Kunpeng Zhang. 2023.
\newblock Multimodal data augmentation for image captioning using diffusion models.
\newblock \emph{arXiv preprint arXiv:2305.01855}.

\bibitem[{Young et~al.(2014)Young, Lai, Hodosh, and Hockenmaier}]{flickr30k}
Peter Young, Alice Lai, Micah Hodosh, and Julia Hockenmaier. 2014.
\newblock From image descriptions to visual denotations: New similarity metrics for semantic inference over event descriptions.
\newblock \emph{TACL}, 2:67--78.

\bibitem[{Zhang et~al.(2022)Zhang, Sugawara, Aizawa, Zhou, Sasano, and Takeda}]{zhang-etal-2022-cross}
Hongkuan Zhang, Saku Sugawara, Akiko Aizawa, Lei Zhou, Ryohei Sasano, and Koichi Takeda. 2022.
\newblock \href {https://doi.org/10.18653/v1/2022.emnlp-main.516} {Cross-modal similarity-based curriculum learning for image captioning}.
\newblock In \emph{Proceedings of the 2022 Conference on Empirical Methods in Natural Language Processing}.

\end{thebibliography}
\bibliographystyle{acl_natbib}

\appendix
\section{Appendix}
\label{sec:appendix}

\subsection{User interface for human study on categorizing text-to-image generation errors}
\label{appendix:user-study}
\begin{figure}[tbh]
        \centering
        \includegraphics[max size=\linewidth]{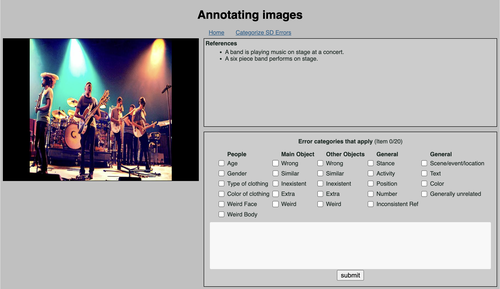}
        \caption{Annotation interface for categorizing SD errors.}
        \label{fig:sd-errors-ui}
    \end{figure}

Our user interface is shown in Figure~\ref{fig:sd-errors-ui}. Annotators were asked to tick boxes of errors that they found in the given synthesized images.

The error categories include:
\begin{itemize}
    \item \textbf{People}: age, gender, type of clothing, color of clothing, weird face, weird body
    \item \textbf{Main object}: wrong, similar, inexistent, extra, weird
    \item \textbf{Other objects}: wrong, similar, inexistent, extra, weird
    \item \textbf{General}: stance, activity, position, number, inconsistent references, scene/event/location, text, color, generally unrelated
    
    \end{itemize}

\subsection{Prompting approaches for text-to-image generation}\label{appendix:prompt}
    \begin{figure}[!tb]
        \centering
        \includegraphics[max size={0.9\linewidth}{0.7\textheight}, trim={2.5cm 1.5cm 4.5cm 2cm},clip]{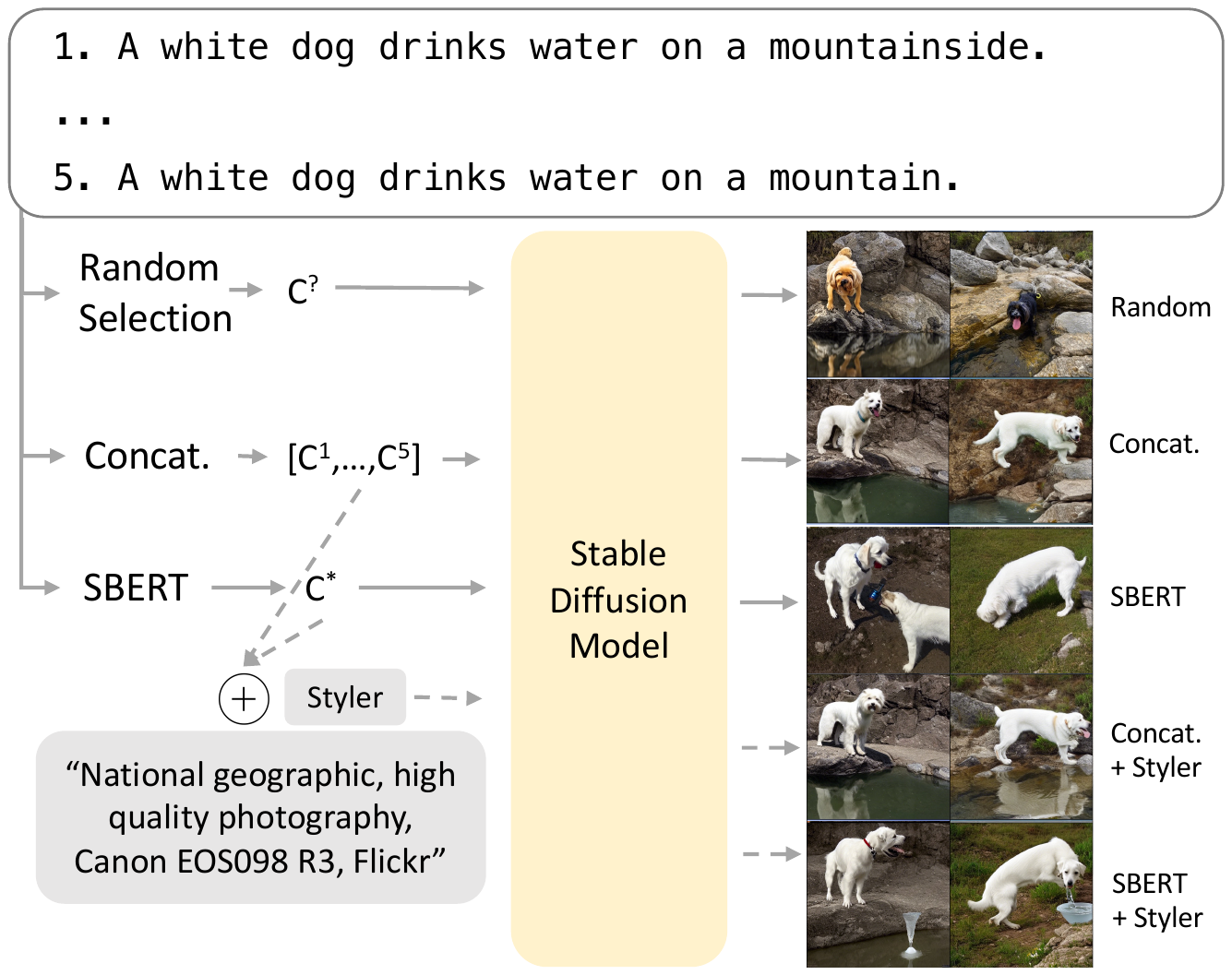}
        \caption{Different prompting strategies for synthetic image generation with text-to-image generation and representative examples. Based on our Round-trip Captioning Evaluation, prompting with the concatenated captions and the styler generates the best images for the task.}
        \label{fig:prompting}
    \end{figure}
    
Figure~\ref{fig:prompting} illustrates the different approaches that we use to prompt the text-to-image generation model. We manually design the styler by inspecting the generated visual examples. 

\subsection{Generating alternative captions with XGLM}
\label{app:xglm}

We follow the prompt template used in \citep{ramos-etal-2023-lmcap} to obtain LM-generated captions, i.e. ``I am an intelligent image captioning bot. Similar images have the following captions: <captions> A creative short caption I can generate to describe this image is: <generation>''. Here we used four ground truth captions as <captions> and the other one in <generation> for a image to build three-shot examples as the prompt. We used the `facebook/xglm-2.9B' model which is available on HuggingFace \cite{wolf2019huggingface}. We set the maximum generation length to 30 tokens with number of beams of 5 to prevent from generating repeated tokens.

 \subsection{Combining multiple curation methods}
    \label{para:combine}
        In our pursuit to assess the efficacy of a hybrid model incorporating multiple curation methods, we experiment on the Flickr30K dataset with BEiT-3 as an initial attempt. For the combining strategy, we selected the two most effective methods on the dataset, namely \textsc{Remove} and \textsc{ReplaceImg}. After each training epoch, we curated the training samples by eliminating one half of the top loss samples while substituting the images of the remaining half. Here we curate on the samples with a loss that exceeded two standard deviations from the mean. Our experiment achieves a CIDEr score of 83.8 and a BLEU4 score of 32.8, surpassing previous single curation performance on the dataset. We believe that the hybrid curation approach would yield greater benefits with more sophisticated combining strategies, which we leave for future work.
        
\subsection{High-loss training samples}
\begin{figure*}
        \centering
        \includegraphics[width=\textwidth]{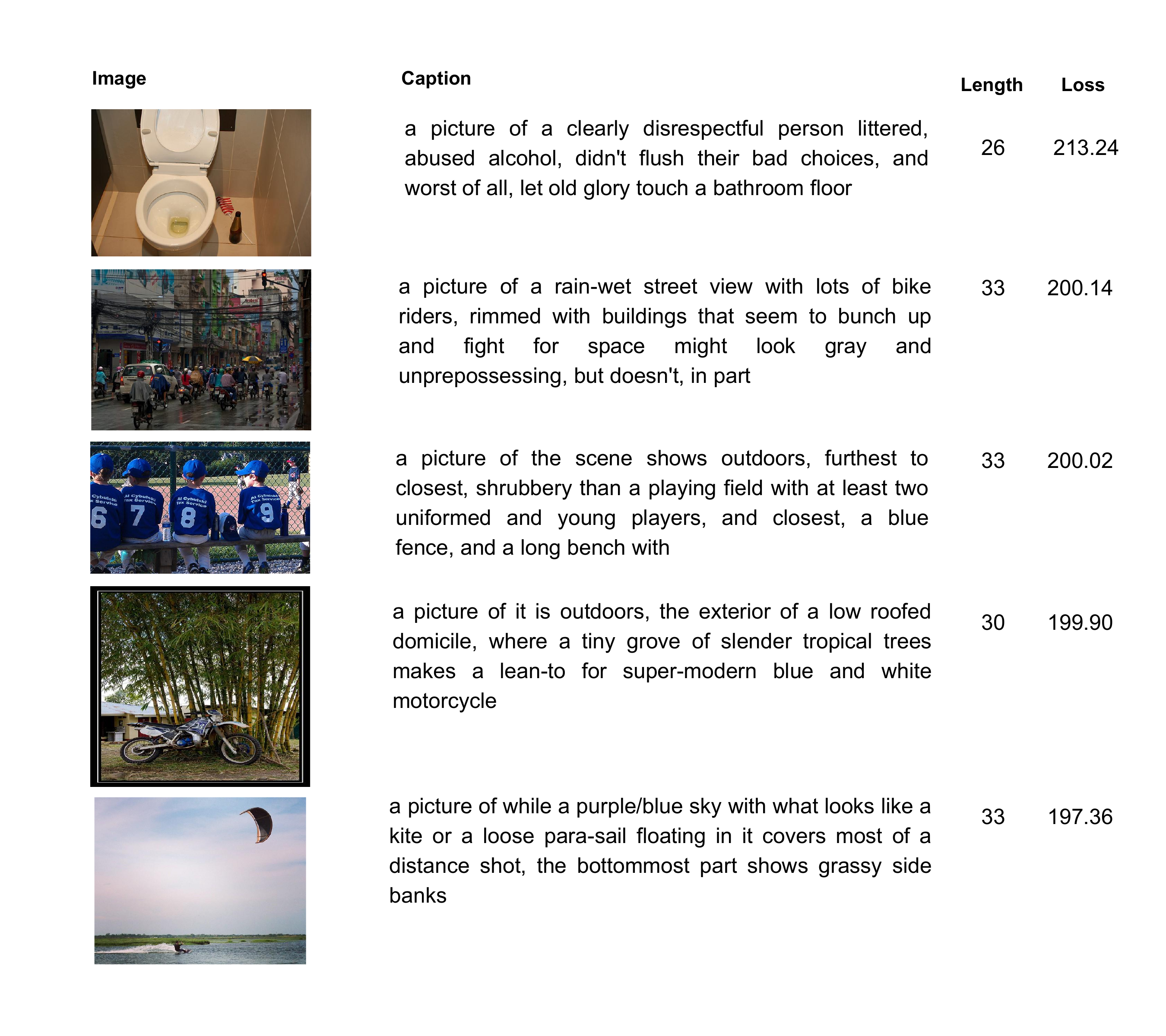}
        \caption{High loss training samples in COCO after the first epoch, ranked by loss in descending order. The top samples all have very long captions around 30 words, compared to the mean of 11 words of the datasets.\label{fig:examples-highloss}}
    \end{figure*}

In Figure~\ref{fig:examples-highloss}, we visualize the high loss training samples in the COCO dataset after the first epoch of finetuning. These samples are target of our curation techniques. Compared to the average caption length of 11 words, the top samples all have very long captions of around 30 words, making it difficult for the model to learn. In the following finetuning epochs, we curate on these samples by either removing the text-image pairs completely (\textsc{Remove}), replacing the caption (\textsc{ReplaceCap}), or replacing the image with a synthesized unseen image (\textsc{ReplaceImg}).

\subsection{Examples of synthesized images}
\begin{figure*}
            \centering
            \includegraphics[width=\textwidth]{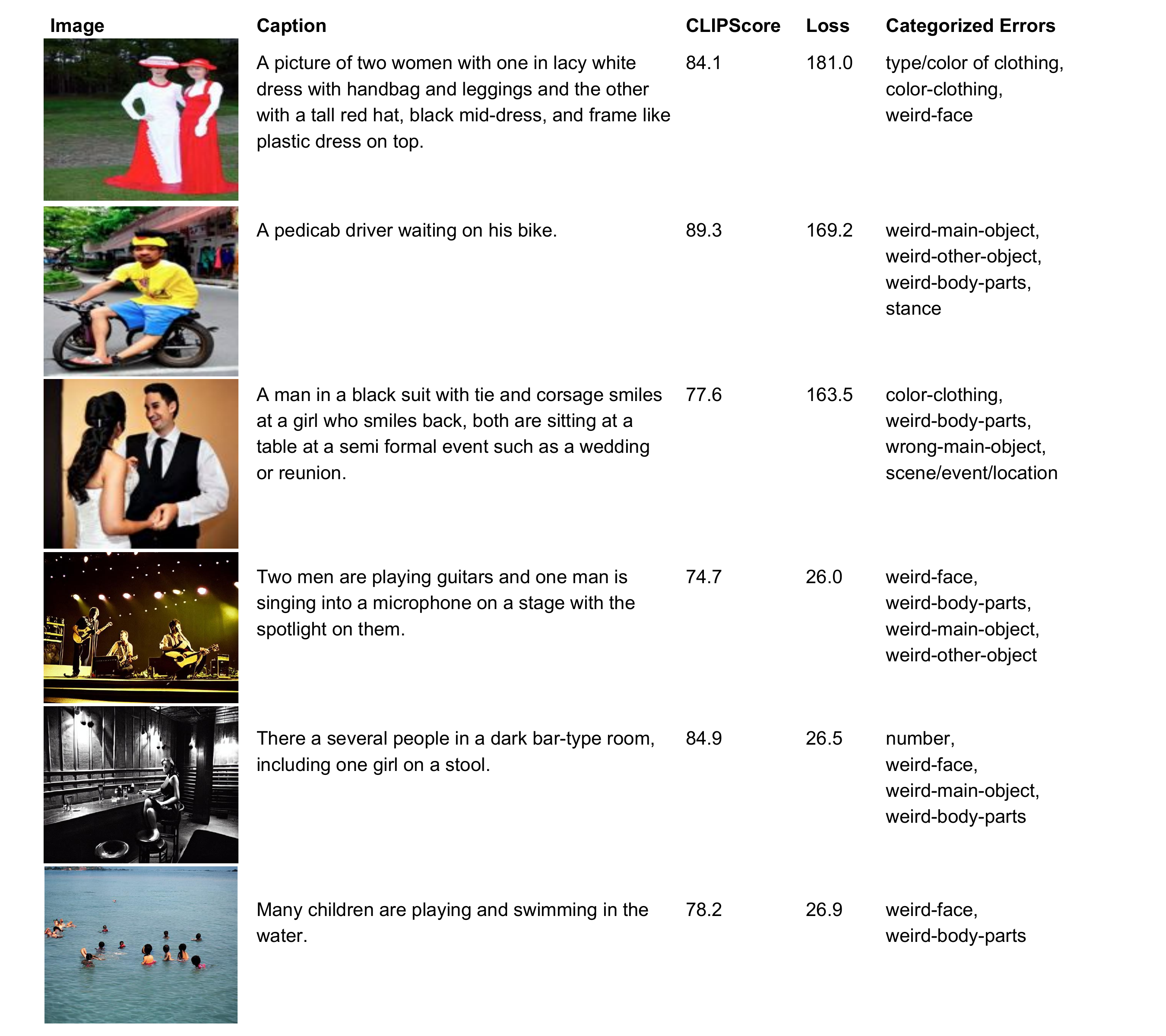}
            \caption{Examples of synthesized images that are of high losses (top) and examples of synthesized images that are of low losses (bottom). Human annotations show that consistent error types have been recognized for the high loss samples while CLIPScore fails to align with human judgement. The low loss synthesized images are visually less complicated than the higher loss ones, but can still often look weird and contain errors in color or objects.\label{fig:examples-loss}}
        \end{figure*}

In Figure~\ref{fig:examples-loss}, we show examples of synthesized images from the text-to-image model that are of high losses and low losses, alongside with the human annotations regarding errors identified from these images.

\end{document}